\documentclass{article}
\usepackage[accepted]{icml2025}
\usepackage{microtype}
\usepackage{graphicx}
\usepackage{subfigure}
\usepackage{booktabs} % for professional tables
\usepackage{bm}
\usepackage{hyperref}

\usepackage{amsmath}
\usepackage{amssymb}
\usepackage{mathtools}
\usepackage{amsthm}

\usepackage[capitalize,noabbrev]{cleveref}

\theoremstyle{plain}

\theoremstyle{definition}

\theoremstyle{remark}

\usepackage[textsize=tiny]{todonotes}

\icmltitlerunning{HyperSAT: Unsupervised Hypergraph Neural Networks for Weighted MaxSAT Problems}
\begin{document}

\twocolumn[
\icmltitle{HyperSAT: Unsupervised Hypergraph Neural Networks for Weighted MaxSAT Problems}

% Unsupervised Hypergraph Neural Networks for Weighted MaxSAT Problems




\begin{icmlauthorlist}
%\icmlauthor{Anonymous Author(s)}{}
\icmlauthor{Qiyue Chen}{yyy,comp}
\icmlauthor{Shaolin Tan}{comp}
\icmlauthor{Suixiang Gao}{yyy,comp}
\icmlauthor{Jinhu L\"{u}}{sch,comp}
\end{icmlauthorlist}

\icmlaffiliation{yyy}{School of Mathematical Sciences, University of Chinese Academy of Science, Beijing, China}
\icmlaffiliation{comp}{Zhongguancun Laboratory, Beijing, China}
\icmlaffiliation{sch}{School of Automation Science and Electrical Engineering, Beihang University, Beijing, China}


\icmlcorrespondingauthor{Shaolin Tan}{shaolintan@hnu.edu.cn}
%\icmlcorrespondingauthor{Firstname2 Lastname2}{first2.last2@www.uk}

%\icmlkeywords{Machine Learning, ICML}

\vskip 0.3in
]



\printAffiliationsAndNotice{}  % leave blank if no need to mention equal contribution

\begin{abstract}
% Graph neural networks (GNNs) have shown promising performance in solving both Boolean satisfiability (SAT) and Maximum Satisfiability (MaxSAT) problems due to their ability to efficiently model and capture the structural dependencies between literals and clauses. 
% Nevertheless, regarding the Weighted MaxSAT problems, GNN methods remain underdeveloped. 
% The challenges come from the non-linear dependency and sensitive objective function owing to the non-uniform distribution of weights across clauses. 
% In this paper, we propose HyperSAT, a novel neural solver that employs an unsupervised hypergraph neural network model, to solve Weighted MaxSAT problems.
% We design a hypergraph representation for Weighted MaxSAT instances, and a cross-attention mechanism along with a shared representation constraint loss function to capture the logical interactions between positive and negative literal nodes in the hypergraph.
% Extensive experiments on different Weighted MaxSAT datasets demonstrate that HyperSAT achieves better performance than state-of-the-art competitors.

Graph neural networks (GNNs) have shown promising performance in solving both Boolean satisfiability (SAT) and Maximum Satisfiability (MaxSAT) problems due to their ability to efficiently model and capture the structural dependencies between literals and clauses. 
However, GNN methods for solving Weighted MaxSAT problems remain underdeveloped.
The challenges arise from the non-linear dependency and sensitive objective function, which are caused by the non-uniform distribution of weights across clauses. 
In this paper, we present HyperSAT, a novel neural approach that employs an unsupervised hypergraph neural network model to solve Weighted MaxSAT problems.
We propose a hypergraph representation for Weighted MaxSAT instances and design a cross-attention mechanism along with a shared representation constraint loss function to capture the logical interactions between positive and negative literal nodes in the hypergraph.
Extensive experiments on various Weighted MaxSAT datasets demonstrate that HyperSAT achieves better performance than state-of-the-art competitors.

\end{abstract}

\section{Introduction}
%The Boolean satisfiability problem, often referred to as SAT, is the first problem proven to be NP-complete in the field of computational complexity(Cook, 1971), and it requires exponential time to solve in the worst case unless P=NP. 许多组合优化问题，例如graph coloring[3], vertex cover[4] and clique detection[5]，都可以规约为sat问题求解。maxsat问题是sat问题的推广，旨在寻找使得布尔逻辑公式中满足子句最大的赋值。sat求解广泛应用于工业领域...

%SAT问题在a，b,c,十分重要，它是第一个被NP。。

The Boolean satisfiability (SAT) problem is a fundamental computational problem that asks whether there exists an assignment of true or false values to a set of variables such that a given propositional Boolean formula evaluates to true. It is the first problem proven to be NP-complete~\cite{cook1971complexity} and has important applications in various fields, including planning~\cite{kautz1992planning}, hardware and software verification~\cite{clarke2001bounded}, and cryptanalysis~\cite{sun2020neurogift}. The Maximum Satisfiability (MaxSAT) problem and its generalization, the Weighted MaxSAT problem, are the variants of SAT with a goal of maximizing the number of satisfied clauses or their weighted sum. By incorporating the notion of prioritizing or weighing different clauses, the Weighted MaxSAT problem provides a more general and flexible version of SAT and MaxSAT to tackle a wide range of real-world problems, such as scheduling~\cite{thornton1998dynamic}, action model learning~\cite{yang2007learning}, computational protein design~\cite{allouche2012computational} and resource allocation~\cite{timm2022synthesis}.

Over the past several decades, SAT problems and their extensions have become central topics in computational logic and theoretical computer science. A wide array of solvers has been developed, with most of them relying on meticulously designed search algorithms~\cite{audemard2018glucose,cai2021deep,biere2022gimsatul}. With the rapid progress of deep learning, there has been a growing shift towards developing learning-based solvers. These learning-based methods aim to effectively extract hidden structures within problem instances and uncover underlying patterns from the data, thus reducing reliance on manually crafted and domain-specific strategies, and providing more adaptable and scalable solvers to tackle large and complex problem instances.

Specifically, the applications of deep learning techniques to SAT problems have led  to two main classes of approaches: end-to-end neural solvers and neural-guided solvers. The end-to-end approaches are typically designed to predict satisfiability or optimal assignments directly from the raw problem input. For example, the groundbreaking work NeuroSAT~\cite{selsam2018learning} introduced an end-to-end framework that predicts satisfiability on random instances using a message passing neural network (MPNN). The study by~\cite{cameron2020predicting} represented SAT instances as permutation-invariant sparse matrices and compared the performances of the exchangeable architecture and MPNN. For the first time, it was demonstrated that end-to-end learning methods could achieve competitive performance in terms of satisfiability prediction. QuerySAT~\cite{ozolins2022goal} proposed a neural SAT solver with a query mechanism that allows the network to make multiple solution trials and get feedback on the loss value towards better solutions.

The neural-guided solvers, on the other hand, integrate neural networks with traditional search frameworks to improve both the efficiency and solution quality of SAT solvers. Along this line, NeuroCore~\cite{selsam2019guiding} leveraged Graph Neural Networks (GNNs) to predict unsatisfiable cores in SAT instances and further applied this prediction to periodically update the activity score of each variable in the Conflict-Driven Clause Learning (CDCL) solver. NLocalSAT~\cite{zhang2020nlocalsat} employed a gated graph convolutional network (GGCN) to guide the initialization of assignments in the Stochastic Local Search (SLS) solver. Recently, NeuroBack~\cite{wang2024neuroback} proposed a new approach by utilizing a graph transformer architecture to make offline neural predictions on backbone variable phases for refining the phase selection heuristic in CDCL solvers.

For the MaxSAT problem, ~\cite{liu2023can} was the pioneering work in exploring the use of GNNs for solving the MaxSAT problem. This work represented the MaxSAT problem as two kinds of factor graphs and studied the capability of typical GNN models in solving the MaxSAT problem from a theoretical perspective. Overall, it can be observed that GNNs have shown promising performance in solving both SAT and MaxSAT problems, however, the extension of these methods to the Weighted MaxSAT problem remains underdeveloped to our best knowledge. The main challenge comes from the uneven weight distribution in the Weighted MaxSAT problem. The addition of weights increases the complexity of the search space and also makes the design of effective heuristics more difficult. The neural model needs to learn how to prioritize high-weight clauses over others. Moreover, since a small change in the assignment could result in a significant change in the weighted sum, the hidden structural patterns within Weighted MaxSAT instances are much harder to learn.

In this work, we consider the problem of designing a neural network solver for Weighted MaxSAT problems. Considering the intrinsic difficulties in learning the non-linear dependency and sensitive objective function in the Weighted MaxSAT problem, we formulate our learning-based framework HyperSAT by integrating hypergraph representation, cross-attention mechanism, and multi-objective loss design. We model the Weighted MaxSAT instance as a hypergraph. The positive and negative literals of a variable are represented as separate nodes, and each clause is represented as a hyperedge, with an associated weight equal to the weight of the clause. The hypergraph convolutional network (HyperGCN) is utilized as the learning model, and a specific cross-attention mechanism is designed to capture the logical interplay between the positive and negative literal node features of the same variable. 
In addition, we design an unsupervised multi-objective loss function to optimize the learning model. Besides the intrinsic optimization objective of the Weighted MaxSAT problem, a shared representation constraint objective is introduced to ensure that the representations of positive and negative literals of a variable are as distinct as possible in the feature space.
We have tested our model on several random Weighted MaxSAT datasets in different settings. The experimental results demonstrate that our model outperforms baseline methods across various datasets, achieving significantly better performance. 
This work provides a new perspective on solving the Weighted MaxSAT problem using learning-based methods, with the hope that the results can offer preliminary knowledge into the capability of neural networks for solving Weighted MaxSAT problems in the future.

In summary, we make the following contributions:
\begin{itemize}
\setlength{\itemsep}{1pt}
\item We propose HyperSAT, an innovative neural approach that uses an unsupervised hypergraph neural network model to solve Weighted MaxSAT problems. This is the first work to predict the solution of the Weighted MaxSAT problem with HyperGCN in an end-to-end fashion.

\item We propose a hypergraph representation for Weighted MaxSAT instances and design a cross-attention mechanism along with a shared representation constraint loss function to capture the logical relationships between positive and negative literal nodes within the hypergraph.
%We propose a hypergraph representation for Weighted MaxSAT instances, incorporating a cross-attention mechanism to model the logical relationships between positive and negative literal nodes within the hypergraph. We design an unsupervised multi-objective loss function that including an intrinsic optimization objective loss and a shared representation constraint loss function.
%We propose a hypergraph representation for Weighted MaxSAT instances, where the logical interactions between positive and negative literal nodes are captured through a cross-attention mechanism. Additionally, a shared representation constraint loss function is introduced to enhance the distinction between these nodes in the feature space.
%We propose a hypergraph representation for Weighted MaxSAT instances, and design a cross-attention mechanism along with a shared representation constraint loss function to capture the logical interplay between positive and negative literal nodes in the hypergraph.

\item We conduct an extensive evaluation of the proposed HyperSAT on multiple Weighted MaxSAT datasets with different distributions. The experimental results demonstrate the superior performance of HyperSAT.
\end{itemize}
%We build a hypergraph nerual nerwork models to solve Weighted MaxSAT problem, which is the first work to predict the solution of Weighted MaxSAT problem with HGNNs in an end-to-end fashion.

%We 提出新的构建cnf formula 的超图表现形式。并且引入cross-attention机制和一个包含共享约束损失函数项来解决正负节点间的logical properties。

%We evaluate the HGNN models on several datasets of random Weighted MaxSAT instances with different distributions, and show that GNNs can achieve good performance and generalization on this task.

\section{Preliminaries}
We now briefly introduce some relevant preliminaries on Weighted MaxSAT and hypergraph neural networks that will be leveraged in later sections. 

\subsection{Weighted MaxSAT}

A Weighted MaxSAT instance is represented by a triple $\varphi=(\mathcal{X},\mathcal{C},\mathbf{w})$. Here, the set of variables is denoted by $\mathcal{X}=\{x_1, x_2, \ldots, x_n\}$, where each $x_i\in\{0,1\}$ is a Boolean variable and $n$ is the total number of variables involved in the instance. The set of clauses is given by $\mathcal{C}=\{C_1, C_2, \ldots, C_m\}$, where each $C_j\in\mathcal{C}$ is a disjunction of literals and $m$ is the number of clauses. The weight vector is given by $\mathbf{w}=(w_1,w_2,\ldots,w_m)$ where $w_j$ represents the weight of $C_j$. Take the clause $C_j\in\mathcal{C}$ for example. Suppose it contains $k_j$ literals. Then, it can be represented by $C_j= l_{j1} \lor l_{j2} \lor \cdots \lor l_{jk_j}$, where each literal $l_{ji}$ is either a variable or the negation of a variable in $\mathcal{X}$. The clause $C_j$ is satisfied if at least one of its literals evaluates to true. The clause $C_j$ is unsatisfied if all its literals evaluate to false.

\begin{figure*}[!t]
\centering
  \includegraphics[width=\linewidth]{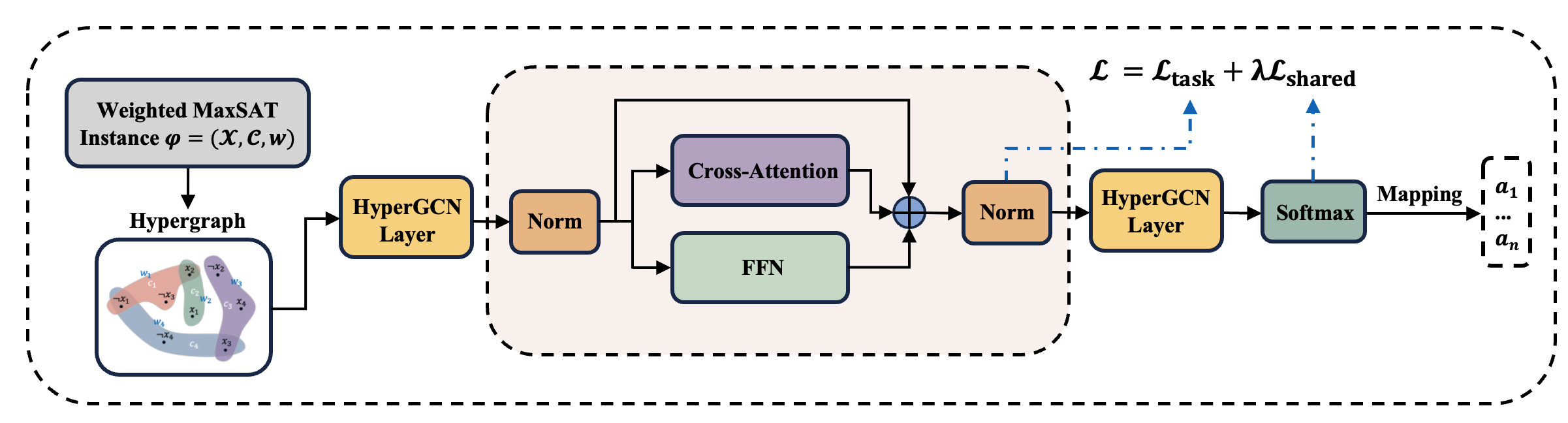}
  \caption{An overview of HyperSAT framework.}
  \label{overview2}
\end{figure*}

An assignment $\bm{A}=\{a_1,a_2,\ldots,a_n\}$ is to assign each variable $x_i\in\mathcal{X}$ with a value $a_i\in\{0,1\}$. Given a Weighted MaxSAT instance $\varphi=(\mathcal{X},\mathcal{C},\mathbf{w})$, the objective is to find an assignment $\bm{A}$ for the Boolean variables $\mathcal{X}$ such that the total weight of satisfied clauses is maximized. The optimization problem is given by
\begin{equation*}
\begin{aligned}
\underset{\bm{A}}{\max} & \quad \sum_{j=1}^m w_j \cdot \mathbf{1}(C_j (\bm{A})) \\
\text{s.t.} & \quad \bm{A}\in\{0,1\}^n.
\end{aligned}
\end{equation*}
Here,
\begin{align*}
\mathbf{1}(C_j(\bm{A})) = 
\begin{cases}
1, & \text{if $C_j$ is satisfied by } \bm{A}, \\
0, & \text{otherwise.}
\end{cases}
\end{align*}

\subsection{Hypergraph Neural Networks}
A hypergraph is defined as $\mathcal{G} = (\mathcal{V}, \mathcal{E},\bm{W})$ which includes a set of nodes $\mathcal{V} = \{v_1, v_2, \cdots, v_{\vert \mathcal{V}\vert }\}$ and a set of hyperedges $\mathcal{E} = \{e_1, e_2, \cdots, e_{\vert \mathcal{E}\vert }\}$. 
$\bm{W}$ is a diagonal matrix of edge weights which assigns a weight to each hyperedge. Each hyperedge $e_j$ is a non-empty subset of nodes (i.e., $\emptyset \neq e_j \subseteq \mathcal{V}$). 
The hypergraph $\mathcal{G}$ can be represented as a $\vert \mathcal{V}\vert  \times \vert \mathcal{E}\vert $ incidence matrix $\bm{H}$, where $\bm{H}_{i,j} = 1$ if $v_i \in e_j$ and $0$ otherwise. For a vertex \( v_i \in \mathcal{V}\), its degree is defined as \( d(v_i) = \sum_{j=1}^{\vert \mathcal{E}\vert }\bm{H}_{i,j} \bm{W}_{j,j} \). For an edge \( e_j \in \mathcal{E} \), its degree is defined as \( \delta(e_j) = \sum_{i=1}^{\vert \mathcal{\mathcal{V}}\vert } \bm{H}_{i,j} \). Additionally, \( \bm{D}_v \) and \( \bm{D}_e \) denote the diagonal matrices of the vertex degrees and the edge degrees, respectively.

Hypergraph Neural Networks (HGNN)~\cite{feng2019hypergraph} perform data representation learning by utilizing a hypergraph structure to capture higher-order dependencies between nodes. In HGNN, hyperedge convolution operations are used to extract features by leveraging the hypergraph Laplacian for spectral convolution. To reduce computational complexity, Chebyshev polynomials are applied to approximate the spectral convolution and avoid the need to compute high-order eigenvectors explicitly. 

Through the hyperedge convolution operation, the $l$-th layer of HGNN can be formulated by 
\begin{align}
\label{eq:1}
\bm{X}^{(l+1)} = \sigma \left( \bm{D}_v^{-1/2} \bm{H} \bm{W} \bm{D}_e^{-1} \bm{H}^\top \bm{D}_v^{-1/2} \bm{X}^{(l)} \bm{\Theta}^{(l)} \right),
\end{align}
where \( \bm{X}^{(l)} \in \mathbb{R}^{\vert\mathcal{V}\vert \times d_l} \) is the signal of the hypergraph at \( l \) layer with \( \vert\mathcal{V}\vert \) nodes and \( d_l \) dimensional features, \( \bm{X}^{(0)} \) is the original signal of the
hypergraph. \( \bm{\Theta}^{(l)} \) is the learnable filter parameter and \( \sigma \) denotes the nonlinear activation function.
%The node feature update for l-ths layer is given by the following formula:

%A hypergraph is defined as $\mathcal{G} = (\mathcal{V}, \mathcal{E})$, which includes a vertex set $\mathcal{V}$, a hyperedge set $\mathcal{E}$. The hypergraph $\mathcal{G}$ of a SAT formula can be represented as a $|\mathcal{V}| \times |\mathcal{E}|$ incidence matrix $\mathcal{H}$, which defined as $(H)_{ij} = 1$ if node $i$ belongs to hyperedge $j$, and $(H)_{ij} = 0$ otherwise.
%Alternatively, $\mathcal{E}$ can be represented with an incidence matrix $\mathbf{H} \in \{0,1\}^{|\mathcal{V}| \times |\mathcal{E}|}$, where $H_{i,j} = 1$ if $v_i \in e_j$ and $0$ otherwise. 

\section{HyperSAT}

In this section, we propose HyperSAT, a neural approach for Weighted MaxSAT problems. \cref{overview2} illustrates the workflow of HyperSAT. It mainly consists of three modules: the hypergraph modeling module, the neural network module, and the probabilistic mapping module. 

Given a Weighted MaxSAT instance, HyperSAT first applies its hypergraph modeling component to represent the instance as a hypergraph. Then, the hypergraph is solved by the neural network module, which is responsible for processing and optimizing the hypergraph. Finally, through a mapping operation, the probabilistic output of the neural network module is mapped into Boolean values, which serves as a solution to the Weighted MaxSAT instance.

Note that the uneven weight of clauses increases the non-linear dependency and sensitivity among variables, the neural network is required to learn how to prioritize the contributions of different clauses. To this end, we propose a specific cross-attention mechanism and introduce an unsupervised multi-objective loss function to capture the logical interplay between the positive and negative literal node representations. These two mechanisms are integrated with the hypergraph convolutional network to form the neural network module of our HyperSAT.

\subsection{Hypergraph Modeling}

Given a Weighted MaxSAT instance $\varphi=(\mathcal{X},\mathcal{C}, \mathbf{w})$, we construct the hypergraph $\mathcal{G} = (\mathcal{V}, \mathcal{E},\bm{W})$ as follows:
% Figure~\ref{hypergraph} illustrates Hypergraph Modeling. 

\begin{itemize}
\item For the construction of the vertex set $\mathcal{V}$, each variable $x_i\in\mathcal{X}$ is represented by two nodes $v_i$ and $v_{n+i}$, which correspond to the positive and negative literals of the variable $x_i$ (i.e., $x_i$ and $\neg x_i$), respectively. % each literal \( l_{ij} \) in the CNF formula corresponds to a node in the hypergraph, so \( V \) consists of the set of all literals: \( V = \{ l_{ij} \mid l_{ij} \in C_i, 1 \leq i \leq m \} \).

\item For the construction of the hyperedge set $\mathcal{E}$, each clause $C_j\in\mathcal{C}$ is represented by a hyperedge \( e_j\in\mathcal{E}\), which connects the nodes corresponding to the literals in $C_j$. Specifically, if clause \( C_j \) consists of the literals \( l_{j1}, l_{j2}, \dots, l_{jk_j} \), then the corresponding hyperedge \( e_j \) connects the nodes corresponding to these literals.

%connects the nodes \( l_{i1}, l_{i2}, \dots, l_{ik_i} \). Formally, \( \mathcal{E} = \{ e_i \mid e_i = \{ l_{i1}, l_{i2}, \dots, l_{ik_i} \}, 1 \leq i \leq m \} \).
\item For the construction of the weight matrix $\bm{W}$, the weight of each hyperedge is equal to the weight of the corresponding clause, i.e., \( \bm{W}_{j,j} = w_j \).
\end{itemize}

\begin{figure}[ht]
\vskip 0.2in
\begin{center}
\centerline{\includegraphics[width=0.9\columnwidth]{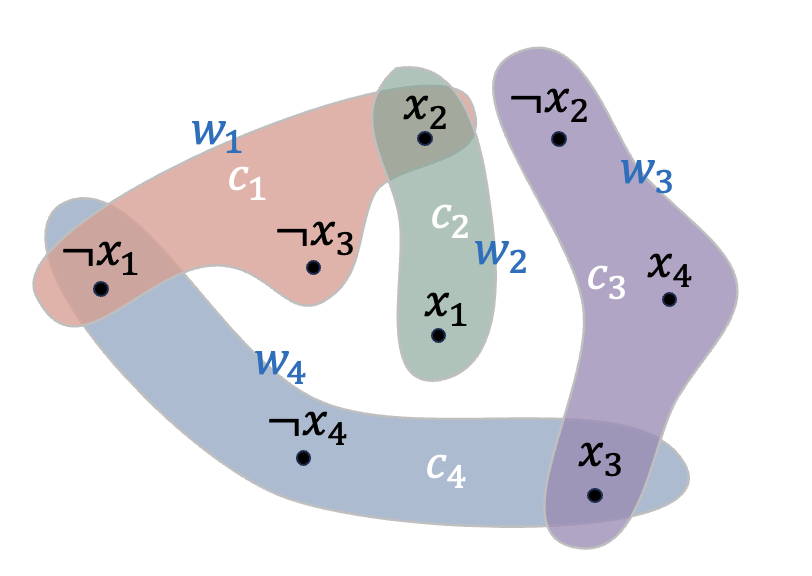}}
\caption{Hypergraph Modeling of a Weighted MaxSAT instance $(\neg x_1 \vee x_2 \vee \neg x_3) \wedge (x_1 \vee x_2) \wedge (\neg x_2 \vee x_3 \vee x_4) \wedge (\neg x_1 \vee x_3 \vee \neg x_4)$ with 8 literals, 4 clauses and weights $\{w_1,w_2,w_3,w_4\}$.}
\label{hypergraph}
\end{center}
\vskip -0.2in
\end{figure}

Through the above modeling process, the Weighted MaxSAT instance can be uniquely represented by a hypergraph, where each literal is a node, and each clause is a hyperedge connecting the literals involved. The weight of each hyperedge is the weight of the corresponding clause in the instance. \cref{hypergraph} gives an illustration of the hypergraph modeling process.

In contrast to existing methods that represent Conjunctive Normal Form (CNF) formulas as factor graphs~\cite{guo2023machine}, which limit their ability to model relationships beyond pairwise connections, we construct Weighted MaxSAT instances as hypergraphs. These hypergraphs can encode higher-order variable dependencies through their degree-free hyperedges, and thus offer more powerful representational capacity and enable more efficient handling of higher-order relationships in combinatorial optimization problems. 
In particular, we treat literals rather than variables  as nodes, which addresses a major issue present in HypOp~\cite{heydaribeni2024distributed}. Since the logical relationships between positive and negative literals are central to the Weighted MaxSAT problem, it is essential to treat them as distinct nodes, rather than merging them into a single node. In addition, we encode the weights of the clauses in the Weighted MaxSAT problem as the weights of the corresponding hyperedges in the hypergraph.

\subsection{Neural Network Architecture}

The neural network architecture of our proposed HyperSAT comprises the HyperGCN, a transformer module with the cross-attention mechanism, and a softmax layer.  
%Section \ref{sec:3.2.1}, 
\subsubsection{Hypergraph Convolutional Networks}
\label{sec:3.2.1}
%再区分下名字
In HyperSAT, we introduce a $T$-layer HyperGCN for message passing among different nodes. The operation at the $l$-th layer of the HyperGCN is formally expressed as follows:
\begin{align}
\bm{L}^{(l+1)'} = \sigma\left(\bm{D}_v^{-\frac{1}{2}} \widetilde{\bm{Q}} \bm{D}_v^{-\frac{1}{2}} \bm{L}^{(l)} \bm{R}^{(l)}\right).
\label{eq:update_rule}
\end{align}
In this formulation, the matrix $\bm{L}^{(l+1)'}$ represents the output of the $l$-th layer; $\bm{D}_v$ is the diagonal matrix of the vertex degrees in the hypergraph; $\bm{L}^{(l)} \in \mathbb{R}^{2n \times d_l}$ is the matrix of node representations at the $l$-th layer, where $d_l$ is the dimension of the $l$-th layer node representations; $\bm{R}^{(l)} \in \mathbb{R}^{d_l \times d_{l+1}}$ is the $l$-th layer learnable weight matrix; and $\widetilde{\bm{Q}}$ is given by
\begin{align}
\widetilde{\bm{Q}} = \bm{H} \widetilde{\bm{D}}_e^{-1} \bm{H}^\top - \text{diag}(\bm{H} \widetilde{\bm{D}}_e^{-1} \bm{H}^\top),
\label{eq:m_definition}
\end{align}
where $\bm{H}$ is the hypergraph incidence matrix, and $\widetilde{\bm{D}}_e = \bm{D}_e - \bm{I}$. Specifically, $\sigma$ denotes the nonlinear activation function and $\bm{L}^{(0)}$ is a learnable input embedding of HyperGCN.

Compared to the updating rule in Eq. (\ref{eq:1}) in traditional HGNN, our HyperGCN focuses the convolutional layer's computation more on the influence of adjacent nodes by removing the diagonal elements of $\widetilde{\bm{Q}}$. This adjustment allows the representation updates of each node to better align with the higher-order relationships of the adjacency structure.

%$D_v$ and $D_e$ denote the diagonal matrices of the edge degrees and the vertex degrees, respectively.

%这个模型和梯度下降有什么好处
%回答为什么要减去对角阵为了解决什么问题！！！！！！！！！！！

%HyperGNNs can function as effective learnable transformation functions, converting input embeddings into original optimization variables. This transformation acts as an acceleration for the solver, offering more efficient optimization paths.
%Nature的

\subsubsection{Cross-Attention Mechanism}
The core of the Weighted MaxSAT problem lies in the logical constraints among variables. Positive and negative literal nodes (e.g., $x$ and $\neg x$) are logically mutually exclusive and strongly correlated, and their relationships directly reflect the underlying structural characteristics of the problem. Considering this, we leverage a cross-attention mechanism to dynamically assign importance weights to the relationships between each pair of complementary literals. 
This mechanism enables the model to adaptively capture critical logical properties such as mutual exclusivity, dependency, and the process of information exchange between positive and negative literal nodes.
Therefore, after updating each node representation in the hypergraph convolution layer, a cross-attention layer is introduced to enhance the representations of positive and negative literal nodes.

Specifically, given the $l$-th layer output of the HyperGCN $\bm{L}^{(l+1)'}$, we divide it into two parts: $\bm{L}^{(l+1)'} = [\bm{L}_{+}^{(l+1)'}, \bm{L}_{-}^{(l+1)'} ]$, where $\bm{L}_{+}^{(l+1)'} \in \mathbb{R}^{n \times d_{l+1}}$ represent the positive literal node representations and  $\bm{L}_{-}^{(l+1)'} \in \mathbb{R}^{n \times d_{l+1}}$ represent the negative literal node representations. The cross-attention mechanism is mathematically represented as follows:
\begin{equation}
\begin{aligned}
\bm{L}_{+}^{(l+1)}  = \text{softmax}\left(\frac{\bm{Q}_{+}^{(l+1)} (\bm{K}_{-}^{(l+1)} )^\top}{\sqrt{d_{l+1}}}\right)\bm{V}_{-}^{(l+1)} ,
\label{eq:positive_attention}
\end{aligned} 
\end{equation}
\begin{equation}
\begin{aligned}
\bm{L}_{-}^{(l+1)}  = \text{softmax}\left(\frac{\bm{Q}_{-}^{(l+1)} (\bm{K}_{+}^{(l+1)} )^\top}{\sqrt{d_{l+1}}}\right)\bm{V}_{+}^{(l+1)} .
\label{eq:positive_attention}
\end{aligned} 
\end{equation}
In this formulation, 
\begin{align*}
\bm{Q}^{(l+1)}_+ = \bm{W}_Q^{(l+1)} \bm{L}_+^{(l+1)'}, \bm{Q}^{(l+1)}_- = \bm{\widetilde{W}}_Q^{(l+1)} \bm{L}_-^{(l+1)'}, \\
\bm{K}^{(l+1)}_+ = \bm{W}_K^{(l+1)} \bm{L}_+^{(l+1)'}, \bm{K}^{(l+1)}_- = \bm{\widetilde{W}}_K^{(l+1)} \bm{L}_-^{(l+1)'}, \\
\bm{V}^{(l+1)}_+ = \bm{W}_V^{(l+1)} \bm{L}_+^{(l+1)'}, 
\bm{V}^{(l+1)}_- = \bm{\widetilde{W}}_V^{(l+1)} \bm{L}_-^{(l+1)'},
\end{align*}
% where $\bm{W}_Q^{(l+1)}$, $\bm{W}_K^{(l+1)}$, $\bm{W}_V^{(l+1)}$ are the learnable query, key and value projection matrices for the positive literal nodes at the $l$-th layer, and $\bm{\widetilde{W}}_Q^{(l+1)}$, $\bm{\widetilde{W}}_K^{(l+1)}$, $\bm{\widetilde{W}}_V^{(l+1)}$ are the corresponding projection matrices for the negative literal nodes at the $l$-th layer. 
where $\bm{W}_Q^{(l+1)}$, $\bm{W}_K^{(l+1)}$, $\bm{W}_V^{(l+1)}$ are the learnable query, key and value projection matrices for the positive literal nodes at the $l$-th layer, and $\bm{\widetilde{W}}_Q^{(l+1)}$, $\bm{\widetilde{W}}_K^{(l+1)}$, $\bm{\widetilde{W}}_V^{(l+1)}$ are the learnable query, key and value projection matrice for the negative literal nodes at the $l$-th layer. 
 
% Finally, the updated node representations are integrated back with the original representations as follows:
% \begin{equation}
% \begin{aligned}
% \bm{L}_{+}^{(l+1)}=\bm{M}_{+}^{(l+1)}+\bm{M}_{+}^{(l+1)'},
% \label{eq:positive_update}
% \end{aligned} 
% \end{equation}
% \begin{equation}
% \begin{aligned}
% \bm{L}_{-}^{(l+1)} =\bm{M}_{-}^{(l+1)} +\bm{M}_{-}^{(l+1)'} .
% \end{aligned} 
% \end{equation}

The final node representation at the $l$-th layer is obtained as $\bm{L}^{(l+1)} =[\bm{L}_{+}^{(l+1)}, \bm{L}_{-}^{(l+1)}] $. The cross-attention mechanism explicitly constructs the interaction between positive and negative literal nodes by enabling each node to incorporate the features of its counterpart. This facilitates a more comprehensive representation of node features for Weighted MaxSAT problems.

Building upon this, the transformer module in HyperSAT incorporates the previously discussed cross-attention layer, along with other components to further enhance the modeling capability. Inspired by the recent advancements in Vision Transformer architecture (ViT-22B~\cite{dehghani2023scaling}), the transformer module in HyperSAT includes a LayerNorm layer~\cite{ba2016layer}, followed by a combination of a cross-attention layer and a Feed-Forward Network (FFN) layer, along with a residual connection and an additional LayerNorm layer. The FFN and cross-attention layers operate in parallel to enhance the efficiency. The residual connection is introduced by adding the outputs from the cross-attention layer, the FFN layer, and the LayerNorm layer. The resulting sum is then passed through another LayerNorm layer to stabilize the representations before proceeding to the next layer. 
%Here, the FFN layer consists of 
%\begin{align}
%FFN(x) = 
%\end{align}

The final layer of the network is a softmax layer. We reshape the iterated \( \bm{L}^{(T)'} \in \mathbb{R}^{2n \times 1} \) into \( \hat{\bm{L}}^{(T)'} \in \mathbb{R}^{n \times 2} \), which serves as the input to the softmax layer. After applying the softmax function, we obtain $\widetilde{\bm{L}} = \text{softmax}(\hat{\bm{L}}^{(T)'})$, which provides the soft node assignments for each literal, interpreted as class probabilities. The probability of assigning a variable to a truth value is recorded by $\bm{Y}=\widetilde{\bm{L}}_{\cdot 1}$, which is the first column of $\widetilde{\bm{L}}$.

%take the first column \( L^{(T)}_{.0} \in \mathbb{R}^{n \times 1} \) which is the embedding of the positive literal nodes

%GNN 生成soft node assignments，可以将其视为类概率
%The softmax function is given by:
%\[
%\text{softmax}(x_i) = \frac{e^{x_i}}{\sum_{j=1}^{2} e^{x_j}} 
%\]
%\quad \text{for each element } x_i \text{ in the vector } x.
%This formula ensures that the output values are scaled between 0 and 1 and sum to 1, which is ideal for classification tasks.
%在网络的末尾，加softmax ，要详细讲2n*1 reshape n*2,数学表示

\subsubsection{Loss Function}
%模仿nature2022 写松弛
%我们提出了一个了无监督的loss，这个loss由两部分组成 primary task loss ,shared representation  constraint loss，放11.
%那句话！！！！！！！！！！！！！！！！！！！！！无监督的优点

%Given a weighted MaxSAT instance $(\mathcal{X},\varphi, \mathbf{w})$ and hypergraph $\mathcal{G} = (\mathcal{V}, \mathcal{E},\mathbf{W})$, we propose a differentiable loss function \( \mathcal{L}(\theta) \), as required for standard backpropagation, by promoting the binary decision variables \( x_i \in \{0,1\} \) to continuous (parametrized) probability parameters \( p_i(\theta) \) with the following relaxation approach

Given a Weighted MaxSAT instance $\varphi=(\mathcal{X},\mathcal{C}, \mathbf{w})$ and its hypergraph $\mathcal{G} = (\mathcal{V}, \mathcal{E},\bm{W})$, we relax the Boolean variables $\mathcal{X}$ into continuous probability parameters \( \bm{Y}(\gamma) \) where $\gamma = (\bm{R}, \bm{L}^{(0)}, \bm{W}_Q, \bm{W}_K, \bm{W}_V, \bm{\widetilde{W}}_Q, \bm{\widetilde{W}}_K, \bm{\widetilde{W}}_V)$ represent the learnable parameters. The relaxation is defined as follows:
\begin{equation}
\mathcal{X}\in \{0,1\}^n \longrightarrow \bm{Y}(\gamma) \in [0,1]^n.
\end{equation}
%which is compatible with standard backpropagation requirements.
%\begin{align}
%x_i \longrightarrow p_i(\theta) \in [0,1].
%\end{align}
%将relax状态空间into连续空间
With this relaxation, we can design a differentiable loss function to optimize the learnable parameters $\gamma$, enabling gradient-based optimization.

In this paper, we propose an unsupervised multi-objective loss function that consists of two components: a primary task loss \( \mathcal{L}_{\text{task}} \) and a shared representation constraint loss \(\mathcal{L}_{\text{shared}} \). 
%This unsupervised loss function eliminates the dependence on large, labeled training datasets typically required by supervised learning approaches. 
This unsupervised loss function obviates the necessity for large, labeled training datasets, which are conventionally indispensable in supervised learning paradigms.
The specific form is as follows:
%To this end, we propose a differentiable unsupervised loss function which using optimization objectives. 
\begin{equation}
\mathcal{L}_{\text{total}} = \mathcal{L}_{\text{task}} + \lambda \mathcal{L}_{\text{shared}},
\label{eq:total_loss}
\end{equation}
where \( \lambda\geq 0 \) is a balancing hyperparameter.

The primary task loss function is the relaxed optimization objective of the Weighted MaxSAT problem, as shown below:
\begin{equation}\label{loss1}
\mathcal{L}_\text{{task}}(\bm{Y}) = \sum_{j=1}^{m} w_j V_j(\bm{Y}),
\end{equation}
where
\begin{equation}
V_j(\bm{Y}) = 1 - \prod_{i \in C_j^+} (1 - y_i) \prod_{i \in C_j^-} y_i.
\label{loss2}
\end{equation}
Here, $C_j^+$ and $C_j^-$ are the index sets of variables appearing in the clause $C_j$ in the positive and negative form, respectively. The term \( V_j(\bm{Y}) \) represents the satisfaction of clause \( C_j \), where a value of 1 indicates that the clause is satisfied, and 0 indicates it is not. The weight \( w_j \) ensures that more important clauses are prioritized during optimization. Minimizing the primary task loss function corresponds to maximizing the weighted sum of satisfied clauses in the Weighted MaxSAT problem. 

%and $i \in c^+$ means the variable $y_i$ occur in the clause $c$ in the positive form. Similarly, $i \in c^-$ means the variable $y_i$ occur in the clause $c$ in the negative form.

%After the penultimate layer of the network, a shared representation constraint is introduced to optimize the representations of positive and negative literal nodes.
The shared representation constraint loss function focuses on the representations of positive and negative literal nodes in the penultimate layer of the HyperSAT network. Its form is given by
\begin{equation}
\mathcal{L}_{\text{shared}} =  \left\| \bm{L}^{(T-1)}_{+} + \bm{L}^{(T-1)}_{-} \right\|_F^2,
\label{eq:shared_loss}
\end{equation}
where $\left\| \cdot \right\|_F$ denotes the Frobenius norm. By applying the shared representation constraint loss function, the network encourages the positive and negative literal nodes to develop distinct feature representations in the learning process. This constraint enhances the separation of the two types of nodes in the feature space.

It should be noted that most neural network-based SAT methods employ supervised learning. However, supervised learning approaches are not particularly well-suited for solving the Weighted MaxSAT problem. On the one hand, a Weighted MaxSAT instance may have multiple satisfying assignments. On the other hand, the non-uniform distribution of weight across the clauses makes it challenging for supervised learning models to detect the hidden structural patterns within the Weighted MaxSAT instance. From this perspective, our proposed unsupervised learning approach offers an  effective alternative approach that enhances generalization capabilities while eliminating the need for labeled instances.

% is no straightforward notion of labels for Weighted MaxSAT instances.

\subsection{Probabilistic Mapping}
% 数学表示，学ROS。再用一句说理的
%After obtaining high-quality solutions to the relaxed continuous problem, we employ a random sampling procedure to derive a discrete solution.
The neural network module takes the hypergraph as input and produces a probability vector $\bm{Y}$ as output, which indicates the probability of each variable taking the truth value. Accordingly, to convert the probability vector $\bm{Y}$ into a Boolean assignment, we transform $\bm{Y}$ into
$n$ Bernoulli distributions, where the $i$-th distribution $\mathcal{B}(y_i)$ is over the discrete values \(\{0, 1\}\). Finally, the node assignments are generated by sampling with the corresponding probability distributions. 

%After passing through the softmax layer, continuous outputs {} are generated for the positive and negative literals of each variable. To convert the continuous outputs of the HyperGNN into integer node assignments, these outputs are transformed into n  Bernoulli distributions over the discrete values \(\{0, 1\}\). The output value of the positive literal represents the probability of the variable being assigned a value of 1, while the output value of the negative literal represents the probability of the variable being assigned a value of 0. Based on this probability distribution, sampling is performed for each variable to generate a discrete set of node assignments.

\section{Experiment}
\subsection{Experimental Settings}
We compare the performance of HyperSAT against GNN-based methods for solving Weighted MaxSAT problems. 
%The source code is available at...

\textbf{Baseline Algorithms.} We compare our proposed HyperSAT against GNN-based methods. The following algorithms are considered as baselines: (i) HypOp~\cite{heydaribeni2024distributed}: an advanced unsupervised learning framework that solves constrained combinatorial optimization problems using hypergraphs; (ii)~\cite{liu2023can}: an innovative supervised GNN-based approach that predicts solutions in an end-to-end manner by transforming CNF formulas into factor graphs. We apply these baselines to solve Weighted MaxSAT problems.

\textbf{Datasets.}
We evaluate the algorithms using random 3-SAT benchmarks from the SATLIB dataset~\cite{hoos2000satlib}. The SATLIB dataset is one of the standard datasets for evaluating SAT solvers. We utilize a range of datasets with varying distributions, from which SAT and UNSAT instances are generated for each distribution. The number of variables in the dataset ranges from $100$ to $250$, while the number of clauses varies between $430$ and $1065$. More details on the datasets are provided in \cref{dataset}. In particular, to generate the required Weighted MaxSAT instances, we assign weights to the clauses in each CNF file within the dataset. These weights are sampled from the integers in the range \( [1, 10] \) uniformly at random. 
\begin{table}[htp]
\caption{The parameters of the datasets.}
\label{dataset}
\vskip 0.15in
\begin{center}
\begin{small}
\begin{sc}
\begin{tabular}{c@{\hskip 5pt}c@{\hskip 5pt}c@{\hskip 5pt}c@{\hskip 5pt}c}
\toprule
Dataset & Instance & Vars & Clauses & Types \\
\midrule
uf100-430 & 1000 & 100 & 430 & SAT \\
uuf100-430 & 1000 & 100 & 430 & UNSAT \\
uf200-860 & 100 & 200 & 860 & SAT \\
uuf200-860 & 100 & 200 & 860 & UNSAT \\
uf250-1065 & 100 & 250 & 1065 & SAT \\
uuf250-1065 & 100 & 250 & 1065 & UNSAT \\
\bottomrule
\end{tabular}
\end{sc}
\end{small}
\end{center}
\vskip -0.1in
\end{table}

\textbf{Model Settings.}
HyperGCN is a two-layer network, with the input dimension set to the square root of the number of variables, and the hidden layer dimension set to half of the square root of the number of variables. The dimension of the cross-attention layer is the square root of the number of variables. The dropout used in the cross-attention layer is $0.1$. The model is optimized using the Adam optimizer~\cite{kingma2014adam} with a learning rate of $7\times10^{-2}$. The balancing hyperparameter of loss function \( \lambda \) is $2\times10^{-3}$. The FFN consists of two linear transformations and a ReLU activation function, with both the input and output dimensions set to the square root of the number of variables, and the hidden layer dimension set to half of the square root of the number of variables. An early stopping strategy is employed, with a tolerance of $10^{-4}$ and a patience of $50$ iterations, terminating training if no improvement is observed over this period. Finally, in the random sampling stage, we perform $5$ independent samplings and return the best solution obtained.

\textbf{Evaluation Configuration.}
All experiments are conducted on an NVIDIA A100-SXM4-40GB GPU, using Python 3.9.30 and PyTorch 1.13.0.

\subsection{Analytical Experiment}
\begin{figure}[ht]
\vskip 0.2in
\begin{center}
\centerline{\includegraphics[width=1\columnwidth]{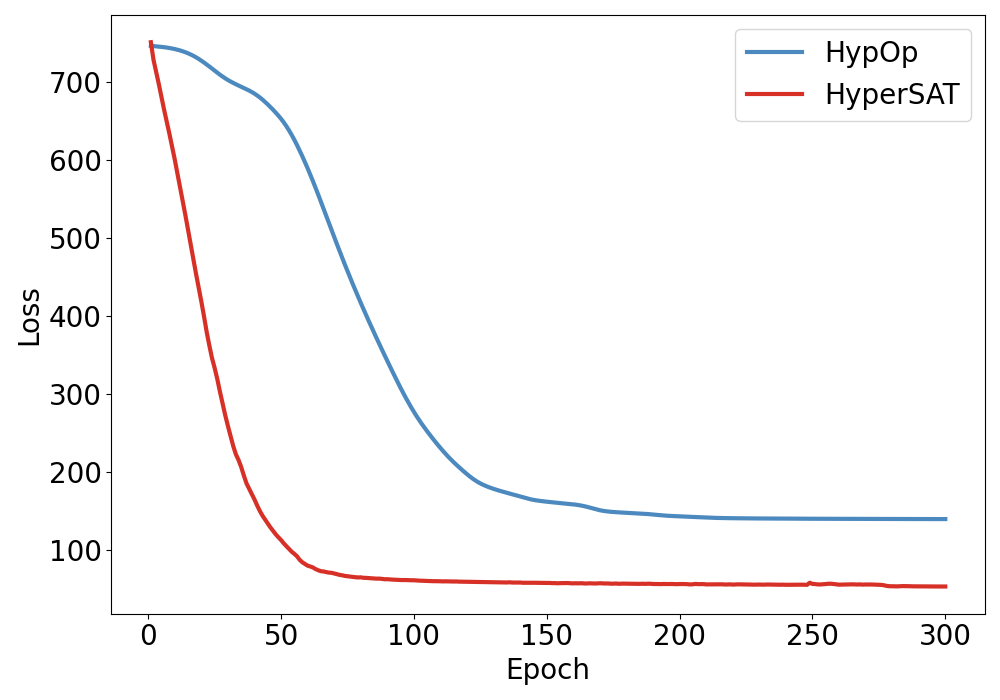}}
\caption{The evolution of loss for HyperSAT and HypOp during an inference process of 300 epochs on the uuf250-1065 dataset.}
\label{curve}
\end{center}
\vskip -0.2in
\end{figure}

We first evaluate the performance of unsupervised methods, HyperSAT and HypOp, with a focus on their convergence. We evaluate the convergence using the primary task loss \( \mathcal{L}_{\text{task}} \) from Eq. (\ref{loss1}), which represents the sum of the weights of unsatisfied clauses. We illustrate the evolution curves of loss for HyperSAT and HypOp on the uuf250-1065 dataset in~\cref{curve} as an example. All models can converge within 300 epochs. Moreover, the loss of HyperSAT is around $52$, while the loss of HypOp is around $139$. We can observe that HyperSAT achieves better performance than HypOp. Specifically, HyperSAT decreases the loss more quickly and achieves a lower loss value. The experimental results demonstrate that HyperSAT can be used in learning to solve Weighted MaxSAT problems.

\subsection{Result}
We evaluate the performance of HyperSAT against baseline algorithms HypOp and~\cite{liu2023can} on various datasets.
The primary evaluation metric is the average weighted sum of unsatisfied clauses. The experiments are conducted on datasets with the number of variables ranging from $100$ to $250$, and the number of clauses varying between $430$ and $1065$.
%我们和。。在table1数据集上进行了。。objective function in （1）的指标进行了比较。
%加的datadet再搬过来 The experiments were conducted across
%The performance of HyperSAT is evaluated on random 3-SAT datasets with corresponding weights, using the average weighted sum of clauses as the primary evaluation metric. 
The results are shown in~\cref{result}.

\begin{table}[htp]
\caption{The average weighted sum of unsatisfied clauses of Weighted MaxSAT problems.}
\label{result}
\vskip 0.15in
\begin{center}
\begin{small}
\begin{sc}
\begin{tabular}{c@{\hskip 5pt}c@{\hskip 5pt}c@{\hskip 5pt}c@{\hskip 5pt}}
\toprule
Dataset & ~\citet{liu2023can} & HypOp &HyperSAT \\
\midrule
uf100-430 & 32.48  & 99.15  & \textbf{15.64} \\
uuf100-430 & 41.65 & 102.44  & \textbf{20.46} \\
uf200-860 & 67.38  & 158.46  & \textbf{28.98} \\
uuf200-860 & 81.68  & 171.34 & \textbf{35.55} \\
uf250-1065 & 79.06  & 170.60 & \textbf{33.24} \\
uuf250-1065 & 100.04  & 182.39 & \textbf{41.64} \\
\bottomrule
\end{tabular}
\end{sc}
\end{small}
\end{center}
\vskip -0.1in
\end{table}

%From the result, HyperSAT efficiently solves all problem instances than all baselines, even for different distribution datasets. 
%The experimental results highlight the superior performance of HyperSAT compared to the baseline algorithms \cite{liu2023can} and HypOp across multiple datasets. 
%As shown in the table, HyperSAT consistently achieves lower values for the average weighted sum of unsatisfied clauses, demonstrating its effectiveness in solving the problem. 
As presented in the table, HyperSAT consistently achieves lower values for the average weighted sum of unsatisfied clauses compared to the baseline algorithms \cite{liu2023can} and HypOp across multiple datasets.
For example, on the uf100-430 dataset, HyperSAT significantly outperforms both baselines with a result of $15.64$. This result represents a substantial improvement over the results of \cite{liu2023can} ($32.48$) and HypOp ($99.15$). 
This trend is observed across all datasets, where HyperSAT always exceeds the performance of the baselines. The average weight of unsatisfied clauses is reduced by approximately $50$\% compared to \cite{liu2023can} and over $80$\% compared to HypOp. 
%This trend is observed across all datasets, where HyperSAT always exceeds the performance of the baselines, with reductions in the average weight of unsatisfied clauses with the ratio of approximately $50$\% compared to \cite{liu2023can} and over $80$\% compared to HypOp.
On the uf200-860 and uuf200-860 datasets, HyperSAT achieves reductions of $56.99$\% and $56.48$\% compared to \cite{liu2023can}, respectively, and reductions of $81.71$\% and $79.25$\% compared to HypOp.
Similarly, on the uf250-1065 and uuf250-1065 datasets, HyperSAT continues to outperform the baselines. The reductions compared to \cite{liu2023can} are $57.96$\% and $58.38$\%, respectively, while the reductions compared to HypOp are $80.52$\% and $77.17$\%.
Importantly, even with the larger datasets, HyperSAT maintains or even enhances its efficacy.
This demonstrates that HyperSAT scales well and performs better.
These results show that HyperSAT consistently provides substantial improvements over both baseline algorithms across a range of datasets, including those with larger problem sizes. As a result, it validates its robustness and effectiveness in reducing the weighted sum of unsatisfied clauses.

%Generalization is an important factor in evaluating the possibility to apply GNN models to solve those harder problems.

\subsection{Ablation Study}
We conduct an ablation study to evaluate the contribution of each key component in our proposed model. 
%By systematically removing or altering individual components, we analyze their effects on performance and provide insights into the importance of each element. 
%缩短
By systematically removing components, we analyze their impact on performance and highlight the importance of each component.
The experiments are designed to isolate the impact of the following components: (i) the hypergraph modeling of literal nodes rather than variable nodes; (ii) the transformer module with the cross-attention mechanism; (iii) the shared representation constraint loss. The results of the ablation study are shown in~\cref{ablation study}.

\begin{table}[t]
\caption{The results of the ablation study. We consider three components: (i) HGM-L: the hypergraph modeling of literal nodes rather than variable nodes; (ii) Transformer: the transformer module with the cross-attention mechanism; (iii) SRCL: the shared representation constraint loss. Specifically, the first row in the table represents the transformation of the Weighted MaxSAT instance into the hypergraph of variables.}
\label{ablation study}
\vskip 0.15in
\begin{center}
\begin{small}
\begin{sc}
\begin{tabular}{lcccr}
\toprule
HGM-L & Transformer & SRCL & Result \\
\midrule
$\times$  & $\times$& $\times$ & 182.39 \\
$\surd$   & $\times$&$\times$ & 86.14 \\
$\surd$  & $\surd$ & $\times$ & 64.08 \\
$\surd$   & $\times$ & $\surd$ &  47.07   \\
$\surd$   & $\surd$ &$\surd$ & 41.64 \\

\bottomrule
\end{tabular}
\end{sc}
\end{small}
\end{center}
\vskip -0.1in
\end{table}

%1-2
\textbf{Effect of Hypergraph Modeling of Literal Nodes:} The performance drops by $52.77$\% when the hypergraph modeling of variable nodes is used instead of literal nodes.
%When the hypergraph modeling of variable nodes is used instead of literal nodes, the performance drops by 52.77\%.
The results demonstrate the importance and superiority of modeling the Weighted MaxSAT instance as a hypergraph with literal nodes.
%which highlights the importance of capturing complex relationships between literals. 
%These results confirm that incorporating the hypergraph modeling of literal nodes plays a crucial role in enhancing the model's effectiveness and efficiency.

%2-3；4-5
\textbf{Effect of Transformer with Cross-Attention:} We disable the transformer module with the cross-attention mechanism to assess its importance. Without this module, the model experiences a performance drop of $11.54$\%. This significantly reduces its ability to effectively capture dependencies across each pair of complementary literals, leading to lower overall accuracy.
%Without the cross-attention mechanism, the model's ability to effectively capture dependencies across different parts of the input is significantly reduced, leading to a lower overall accuracy. 

\textbf{Effect of Share Representation Constraint Loss:} We further remove the shared representation constraint loss and optimize the model with the primary task loss function to investigate the role of the unsupervised multi-objective loss. The results show that without the shared representation constraint loss, the model achieves a performance of $64.08$. This performance is lower than that of the original configuration. Therefore, it highlights the importance of the shared representation constraint loss, which encourages the positive and negative literal nodes to develop distinct feature representations.

%Removing the shared constraints loss has a notable impact on performance, resulting in a performance of X. This highlights the importance of the shared constraints loss, which encourages the positive and negative literal nodes to develop distinct feature representations. Without this module, the model struggles to maintain this separation, leading to a decrease in overall performance.

%We further investigate the role of the unsupervised multi-objective loss by removing the shared constraints loss and optimize the model with a single objective loss function instead. The results demonstrate that removing the shared constraints loss and relying solely on a single objective loss leads to a noticeable drop in performance.   
%The unsupervised multi-objective loss which including the primary task loss and shared constraints loss enhances performance and robustness, ultimately producing a more effective solution compared to the single objective loss approach.

In summary, the full model consistently outperforms the ablated versions, demonstrating the synergistic effect of integrating the hypergraph modeling of literal nodes, cross-attention mechanism, and shared representation constraint loss design.

\section{Conclusion}

In this work, we propose HyperSAT, a novel neural approach for Weighted MaxSAT problems, addressing the challenges associated with learning the complex non-linear dependencies and sensitive objective function of this NP-hard problem. By modeling Weighted MaxSAT instances as hypergraphs, a hypergraph convolutional network is introduced as the learning model, coupled with a cross-attention mechanism to capture the logical relationships between positive and negative literals. The proposed framework incorporates an unsupervised multi-objective loss design, which not only optimizes the intrinsic objectives of the Weighted MaxSAT problem but also ensures that the representations of positive and negative literals remain distinct in the feature space.

Extensive experiments demonstrate the potential of HyperSAT. The experimental results show that HyperSAT is effective in solving Weighted MaxSAT instances and outperforms state-of-the-art competitors in terms of performance. This work offers a fresh perspective on solving the Weighted MaxSAT problem through learning-based methods. Future work will focus on further integrating the model with heuristic solvers, improving the design of well-crafted solvers, and exploring its applications in other complex combinatorial problems.

\section*{Impact Statement}

This paper presents work whose goal is to advance the field of Machine Learning. There are many potential societal consequences of our work, none which we feel must be specifically highlighted here.


\bibliography{example_paper}
\bibliographystyle{icml2025}

%%%%%%%%%%%%%%%%%%%%%%%%%%%%%%%%%%%%%%%%%%%%%%%%%%%%%%%%%%%%%%%%%%%%%%%%%%%%%%%
%%%%%%%%%%%%%%%%%%%%%%%%%%%%%%%%%%%%%%%%%%%%%%%%%%%%%%%%%%%%%%%%%%%%%%%%%%%%%%%
% APPENDIX
%%%%%%%%%%%%%%%%%%%%%%%%%%%%%%%%%%%%%%%%%%%%%%%%%%%%%%%%%%%%%%%%%%%%%%%%%%%%%%%
%%%%%%%%%%%%%%%%%%%%%%%%%%%%%%%%%%%%%%%%%%%%%%%%%%%%%%%%%%%%%%%%%%%%%%%%%%%%%%%
%\newpage
%\appendix
%\onecolumn
%\section{You \emph{can} have an appendix here.}

%You can have as much text here as you want. The main body must be at most $8$ pages long.
%For the final version, one more page can be added.
%If you want, you can use an appendix like this one.  

%The $\mathtt{\backslash onecolumn}$ command above can be kept in place if you prefer a one-column appendix, or can be removed if you prefer a two-column appendix.  Apart from this possible change, the style (font size, spacing, margins, page numbering, etc.) should be kept the same as the main body.
%%%%%%%%%%%%%%%%%%%%%%%%%%%%%%%%%%%%%%%%%%%%%%%%%%%%%%%%%%%%%%%%%%%%%%%%%%%%%%%
%%%%%%%%%%%%%%%%%%%%%%%%%%%%%%%%%%%%%%%%%%%%%%%%%%%%%%%%%%%%%%%%%%%%%%%%%%%%%%%

\end{document}